\documentclass[a4paper, 10pt, conference]{ieeeconf}      
\usepackage{FG2025}
\usepackage{marvosym} 

\IEEEoverridecommandlockouts                              
\def\FGPaperID{260} 

\title{\LARGE \bf
AutoMR: A Universal Time Series Motion Recognition Pipeline
}

\author{\parbox{16cm}{\centering
    {\large Likun Zhang$^{1,2}$ \texttt{<likun.zhang@xintelligencelabs.ac>}, \\Sicheng Yang$^1$ \texttt{<sicheng.yang@xintelligencelabs.ac>}, \\Zhuo Wang$^1$ \texttt{<zhuo.wang@xintelligencelabs.ac>}, \\Haining Liang$^{3}$ \texttt{<hainingliang@hkust-gz.edu.cn>}, \\Junxiao Shen$^{4, \textrm{\Letter}}$\thanks{\footnotesize Corresponding author} \texttt{<junxiao.shen@bristol.ac.uk>}}\\
    \vspace{0.5cm} 
    {\normalsize
    $^1$ X-Intelligence Labs 
    $^2$ University of California, Berkeley 
    $^3$ HKUST (Guangzhou) 
    $^4$ University of Bristol \\
    }
}}

\usepackage{graphicx}
\usepackage{amsmath}
\usepackage{booktabs}
\usepackage{hyperref}

\FGfinalcopy 

\begin{document}

\ifFGfinal
\thispagestyle{empty}
\pagestyle{empty}
\else
\author{Anonymous FG2025 submission\\ Paper ID \FGPaperID \\}
\pagestyle{plain}
\fi
\maketitle

\begin{abstract}

In this paper, we present an end-to-end automated motion recognition (AutoMR) pipeline designed for multimodal datasets. The proposed framework seamlessly integrates data preprocessing, model training, hyperparameter tuning, and evaluation, enabling robust performance across diverse scenarios. Our approach addresses two primary challenges: 1) variability in sensor data formats and parameters across datasets, which traditionally requires task-specific machine learning implementations, and 2) the complexity and time consumption of hyperparameter tuning for optimal model performance. Our library features an all-in-one solution incorporating QuartzNet as the core model, automated hyperparameter tuning, and comprehensive metrics tracking. Extensive experiments demonstrate its effectiveness on 10 diverse datasets, and most achieve state-of-the-art performance. This work lays a solid foundation for deploying motion-capture solutions across varied real-world applications.

\end{abstract}


\section{INTRODUCTION}

\begin{figure}
    \centering
    \includegraphics[width=1\linewidth]{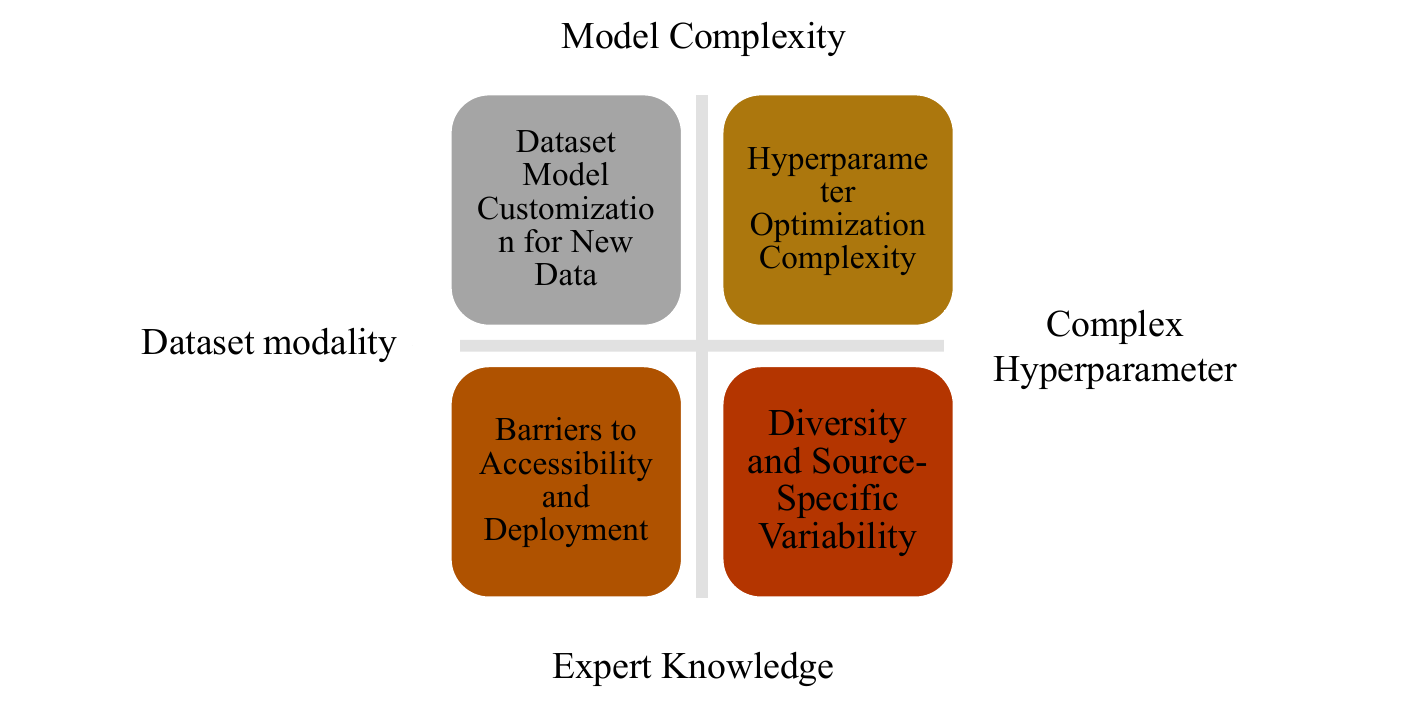}
    \caption{Key challenges in motion recognition: (1) Dataset variability across different sensor types increases preprocessing complexity, (2) Model adaptation requires dataset-specific optimizations, leading to high computational costs, (3) Hyperparameter tuning demands expertise and computational resources, and (4) Deployment barriers limit accessibility for non-specialists. These challenges highlight the need for a scalable and automated framework.}
    \label{fig:challenges}
\end{figure}

Gesture and motion analysis is crucial in fields such as human-computer interaction~\cite{9666999}, healthcare, and robotics. The growing availability of multimodal datasets from wearable sensors \cite{shen2024ringgesture}, cameras, and motion capture systems has driven the need for scalable and adaptable gesture recognition solutions. However, leveraging these datasets effectively presents several challenges, as illustrated in Figure~\ref{fig:challenges}.

One major challenge is the variability in dataset formats, sampling rates, and noise levels. Data collected from different sensors, such as IMUs, skeletal motion capture, or video-based tracking, often require dataset-specific preprocessing pipelines, increasing complexity and limiting scalability. Additionally, gesture recognition models typically need to be tailored to specific datasets or application scenarios. The absence of a unified framework results in repetitive model design and optimization efforts, leading to increased computational costs and resource demands.

Another obstacle is the complexity of hyperparameter tuning. Selecting appropriate learning rates, batch sizes, and network architectures is critical for achieving high performance but often requires domain expertise and substantial computational resources. This challenge is particularly pronounced for non-experts who lack experience in fine-tuning deep learning models. Furthermore, many existing gesture recognition frameworks require extensive customization and expert knowledge \cite{jiang2021emerging}, making them less accessible to non-specialists. Simplifying the deployment process is essential to enable broader adoption across various fields \cite{xu2023xair}.

To address these challenges, this paper introduces \textit{AutoMR}, a unified framework that automates the motion recognition pipeline, encompassing data preprocessing, model training, and hyperparameter tuning. AutoMR standardizes datasets, eliminates the need for manual preprocessing, and enhances performance through automated tuning. Its compatibility with multiple datasets and sensor modalities ensures scalability and facilitates deployment. Evaluated on ten benchmark datasets, AutoMR demonstrates competitive performance against state-of-the-art models. By reducing manual effort and computational costs, the framework streamlines model development and improves accessibility, making motion recognition more efficient for researchers and practitioners.
To fully assess the effectiveness and reproducibility of our method, We encourage readers to review the provided code: \url{https://github.com/X-Intelligence-Labs/AutoMR}. This code demonstrates our state-of-the-art (SOTA) results achieved on eight datasets and showcases the fully automatic and adaptive nature of our model without any parameter tuning. More details are available on the website: \url{https://x-intelligence-labs.github.io/AutoMR_website/}.

\section{RELATED WORK}

Existing research on gesture and motion data analysis spans task-specific models~\cite{liang2024mask}, AutoMR frameworks~\cite{damdoo2020adaptive}, and hyperparameter tuning tools~\cite{nayak2021hyper}. While each of these areas has made significant contributions, they also present certain limitations when applied to multimodal datasets. 

\subsection{Task-Specific Models}
Task-specific models are widely used in gesture and motion recognition. For instance, DeepConvLSTM combines convolutional layers and LSTMs to capture spatiotemporal patterns, showing strong performance on UCI-HAR~\cite{khatun2022deep}.
Similarly, Temporal Convolutional Networks (TCNs) use dilated convolutions to model long-range dependencies in time-series data~\cite{airlangga2024performance}, demonstrating success on OPPORTUNITY~\cite{al2024tcn} and other multimodal datasets. However, these methods are typically tailored to specific modalities (e.g., IMU or skeletal motion) and often lack scalability to new data types such as sEMG signals or unsegmented motion capture data.

\subsection{Auto Training Frameworks for Time-Series Data}
Auto training frameworks, such as Auto-sklearn and H2O AutoGR~\cite{9534091}, offer automation in model selection and hyperparameter tuning. These tools reduce the expertise required to build machine learning pipelines and have demonstrated potential in time-series classification.
However, these frameworks are not tailored for multimodal datasets and often require extensive customization for preprocessing and handling heterogeneous sensor data. For example, TPOT’s limited modality-specific pipelines restrict its applicability to datasets like LMDHG, where skeletal motion data demands complex preprocessing and feature extraction~\cite{gijsbers2018layered}.

\subsection{Hyperparameter Tuning Tools}

Frameworks such as Optuna and Hyperopt are commonly used for optimizing hyperparameters~\cite{shekhar2021comparative}. These tools use algorithms like Bayesian optimization to efficiently explore the search space, and they have been applied to datasets like MHEALTH~\cite{talaat2022effective} and OPPORTUNITY~\cite{ozcan2020human} for tuning hyperparameters such as learning rates and network depths. 
However, they are not integrated into end-to-end workflows and often require significant manual setup for data preprocessing, model training, and evaluation.


\addtolength{\textheight}{-3cm}   

\section{METHOD}

\begin{figure}
    \centering
    \includegraphics[width=1\linewidth]{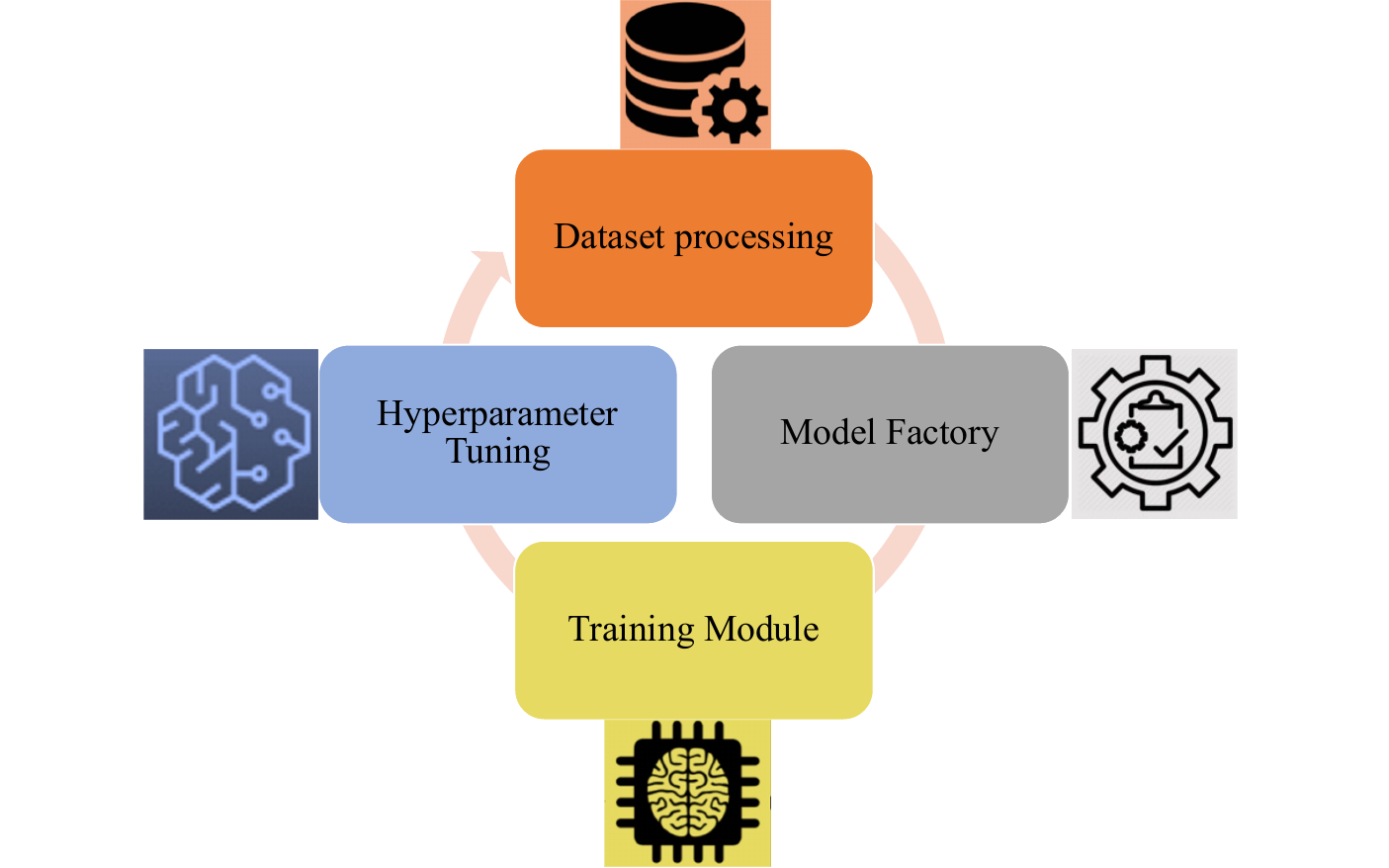}
    \caption{AutoMR end-to-end architecture, illustrating the complete workflow from data preprocessing to model training and hyperparameter tuning. The framework standardizes diverse datasets, selects optimal model configurations, and ensures efficient training and deployment for motion recognition across different sensor modalities.}
    \label{figure}
\end{figure}

\begin{figure}
    \centering
    \includegraphics[width=1\linewidth]{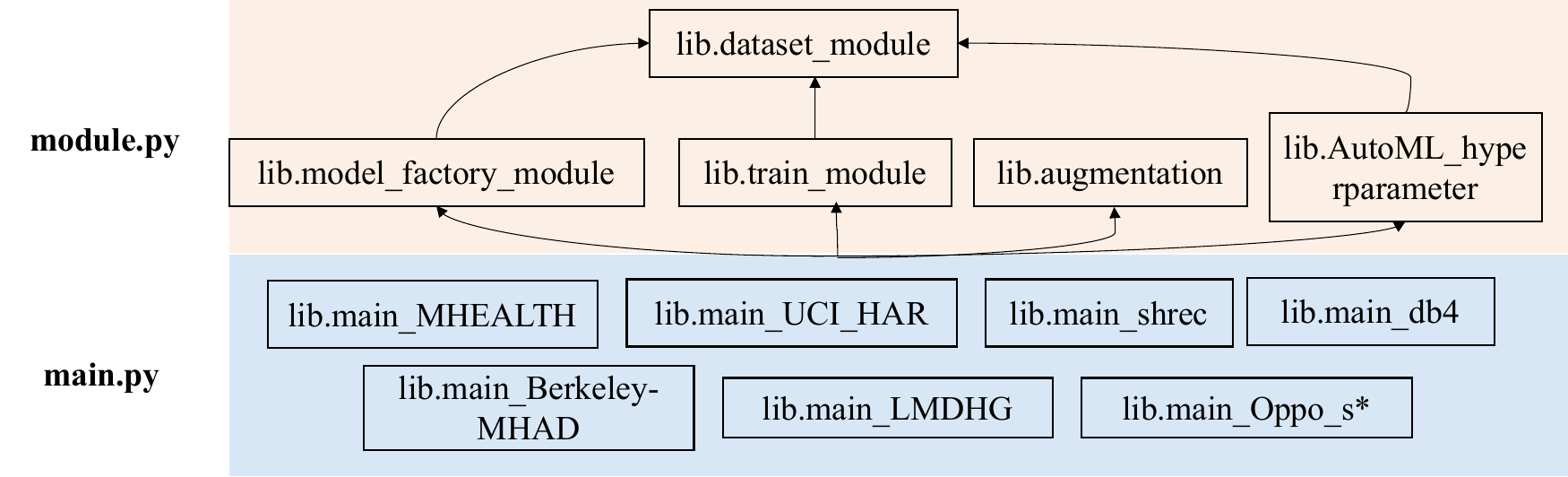}
    \caption{AutoMR's modular and hierarchical architecture, illustrating the interaction between core modules and dataset-specific training scripts. The upper layer consists of fundamental components for dataset preprocessing, augmentation, model selection, training, and hyperparameter tuning, ensuring a standardized and optimized workflow. The lower layer contains dataset-specific execution scripts that utilize these core modules for model training on individual datasets, enabling efficient adaptation across diverse sensor modalities.}
    \label{dependencies}
\end{figure}

\subsection{Architecture Overview}





AutoMR provides a unified solution for motion recognition, addressing challenges such as dataset diversity, model scalability, and hyperparameter optimization, as illustrated in Figure~\ref{figure}.
The core layer includes modules responsible for dataset preprocessing, augmentation, training management, model selection, and hyperparameter tuning (Figure~\ref{dependencies}). The dataset module standardizes multimodal input formats, ensuring consistency across different datasets, while the augmentation module introduces transformations to improve generalization. The training module orchestrates model learning, incorporating configurations from the model factory module and optimizing performance through the AutoML hyperparameter module.

The dataset-specific layer consists of training scripts such as main MHEALTH, main UCI-HAR, and so on, which interact with the core modules to preprocess data, configure models, and execute training. This structured approach minimizes manual adaptation while ensuring scalability, allowing AutoMR to generalize across diverse datasets and facilitate efficient motion recognition.

\subsection{Data Processing and Model Factory}

\begin{table*}[ht]
    \centering
    \caption{description and processing}
    \label{tab:datasets}
    \begin{tabular}{l c c p{9cm}}
        \hline
        \textbf{Dataset} & \textbf{Sequences} & \textbf{Batch Size} & \textbf{Processing and Description} \\
        \hline
        SHREC2021 & 180 & 128 & Contains 3–5 gestures per sequence with 18 gesture classes (static and dynamic). Split into 108 training and 16 test examples per class \cite{10.1016/j.cag.2021.07.007}. \\
        MHEALTH & 10 participants & 256 & Records body motion and vital signs via chest, wrist, and ankle sensors. Segmented into 25-timestamp windows with 50\% overlap \cite{banos2015framework, banos2014mHealthDroid}. \\
        UCI-HAR & 30 participants & 32 & Smartphone-based accelerometer and gyroscope signals. Preprocessed and segmented into 2.56s sliding windows with 50\% overlap, generating 128 readings per window \cite{8975649, pang2021stacked}. \\
        DB4 & 10 participants & 256 & sEMG and kinematic data for 52 hand gestures plus rest. Segmented into 260ms windows with 235ms overlap. Data split in a 4:1 ratio and stored as .txt files \cite{josephs2020semg, 8630679}. \\
        Berkeley-MHAD & 12 participants & 256 & Includes 11 dynamic actions repeated 5 times, totaling 660 sequences. Includes background recordings and T-poses for skeleton extraction \cite{6474999}. \\
        LMDHG & Unsegmented & 64 & Hand gesture sequences annotated with action labels. Skeleton data projected to 2D for visualization. Training/testing split is 4:1 \cite{boulahia2016hif3d, boulahia2017dynamic}. \\
        OPPORTUNITY & Wearable sensors & 256 & Captures human activities using wearable, object, and ambient sensors. Data segmented into 15-frame windows with 7-frame overlap. Augmented to address class imbalance \cite{chavarriaga2013opportunity}. \\
        \hline
    \end{tabular}
\end{table*}

To ensure dataset consistency, we define structured formats that include modality specifications, gesture labels, and sampling parameters. This modular approach supports ten datasets—Shrec2021, MHEALTH, UCI-HAR, DB4, Berkeley-MHAD, LMDHG, and four subsets of the OPPORTUNITY dataset (see TABLE~\ref{tab:datasets})—and aligns diverse data modalities for compatibility with subsequent model training.



In addition, our model factory dynamically selects, configures, and instantiates models for gesture recognition. We choose QuartzNet as our primary model due to its efficient processing of sequential data~\cite{shen2024towards}. QuartzNet employs depthwise separable convolutions to reduce parameters while preserving feature extraction, and its residual connections improve gradient flow for deeper architectures. Varying kernel sizes and dilation rates enable multi-scale temporal pattern recognition~\cite{kriman2020quartznet}. To accommodate datasets of varying complexity, we also employ QuartzNetLarge, which allows adjustments in block structure, input channels, and head channels~\cite{nikam2022robust}.


\subsection{Training Module}

The training module, built on TorchTNT, optimizes QuartzNet by automating metric tracking and incorporating techniques such as gradient clipping, anomaly detection, and dynamic learning rate scheduling. It supports multi-class classification using cross-entropy loss and evaluates performance using metrics like accuracy, F1 score, and confusion matrices. 
Our module includes checkpointing mechanisms to prevent data loss and ensure reproducibility by saving the best-performing model parameters, and we use TensorBoard for real-time logging to facilitate monitoring and debugging.

\subsection{Hyperparameter Tuning}

AutoMR automates hyperparameter tuning using SMAC~\cite{10.1145/3377929.3389999} to dynamically optimize performance. The configuration space adapts to each dataset by tuning general parameters (learning rates, weight decay, dropout rates, batch sizes) and QuartzNet-specific settings (number of blocks, cells per block, input channels, kernel sizes), with the ConfigSpace library managing these ranges to balance cost and efficiency.


The process begins with the ModelHyperOptimizer defining parameter ranges and initializing models. We then train for a fixed number of epochs using standard techniques, evaluating performance via accuracy, F1 score, and loss. SMAC iteratively refines the search, storing the best configurations for reuse unless re-optimization is required. For ablation studies, we also perform manual tuning with batch sizes of 32, 64, 128, or 256 to assess their impact on performance.

\section{EXPERIMENTAL RESULTS ANALYSIS}        

\subsection{Comparison with State-of-the-Art Methods}

\begin{figure}[tp]
    \centering
    \includegraphics[width=1\linewidth]{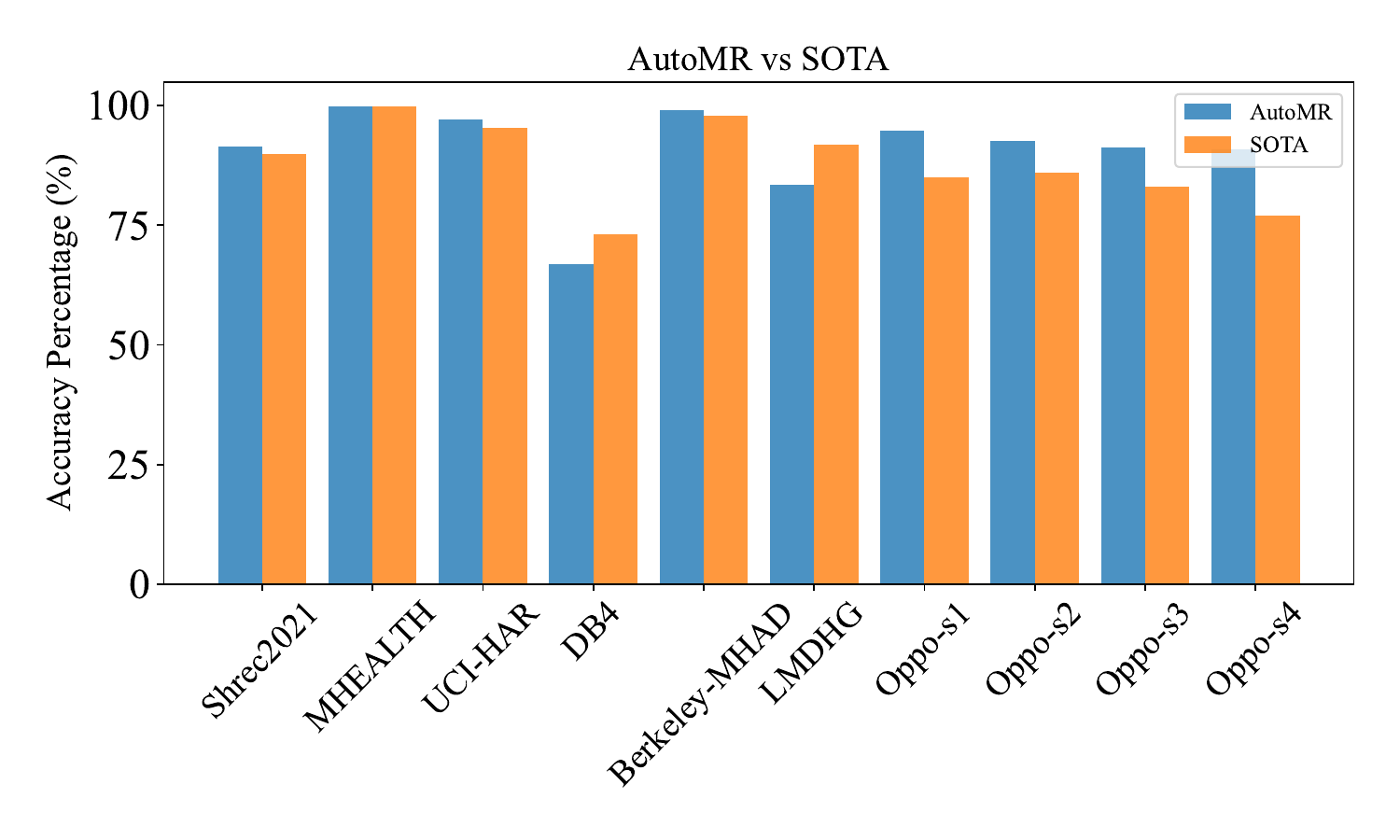}
    \caption{Overall accuracy comparison between AutoMR and SOTA models across ten datasets. AutoMR achieves superior performance on eight datasets, highlighting its effectiveness in generalizing across diverse gesture recognition tasks.}
    \label{figure3}
\end{figure}

Table~\ref{table_performance} and Figure~\ref{figure3} compare the performance of AutoMR with existing state-of-the-art (SOTA) methods across ten benchmark datasets. AutoMR achieves the highest accuracy on eight of the ten datasets, demonstrating its generalizability and adaptability across different sensor modalities and gesture recognition tasks. Notably, AutoMR significantly outperforms previous models on datasets such as OPPORTUNITY, where it achieves over 5\% higher accuracy compared to prior SOTA models. However, performance on DB4 and LMDHG is slightly lower than SOTA, indicating potential areas for future improvement, particularly in handling highly noisy or unstructured motion data.

\begin{table}[t]
    \centering
    \caption{Comparison of AutoMR performance with state-of-the-art (SOTA) models on ten datasets. AutoMR outperforms SOTA methods on eight datasets, demonstrating its robustness and adaptability across different sensor modalities.}
    \label{table_performance}
    \begin{tabular}{ccc}
        \toprule
        \textbf{Dataset} & \textbf{AutoMR Accuracy (\%)} & \textbf{SOTA Accuracy (\%)} \\
        \midrule
        Shrec2021~\cite{10.1016/j.cag.2021.07.007}        & 91.48                  & 89.93 \\
        MHEALTH~\cite{praba2023harnet}          & 99.81                  & 99.80 \\
        UCI-HAR~\cite{8975649}           & 97.05                  & 95.25\\
        DB4~\cite{josephs2020semg}              & 66.86                  & 73.00 \\
        Berkeley-MHAD~\cite{10719991}    & 98.98                  & 97.91 \\
        LMDHG~\cite{boulahia2017dynamic}            & 83.38                  & 91.83 \\
        Oppo-s1~\cite{chavarriaga2013opportunity}          & 94.63                  & 85.00 \\
        Oppo-s2~\cite{chavarriaga2013opportunity}          & 92.59                  & 86.00 \\
        Oppo-s3~\cite{chavarriaga2013opportunity}          & 91.16                  & 83.00 \\
        Oppo-s4~\cite{chavarriaga2013opportunity}          & 90.89                  & 77.00 \\
        \bottomrule
    \end{tabular}
\end{table}

\subsection{Ablation Study}

\begin{table}[!tp]
    \centering
    \setlength{\tabcolsep}{3pt} 
    \renewcommand{\arraystretch}{0.9} 
    \caption{Ablation study comparing automatic and manual hyperparameter tuning across four key metrics: accuracy, precision, recall, and F1-score. The results show that AutoMR’s automatic tuning achieves performance comparable to manual tuning, demonstrating its reliability for real-world applications.}
    \label{table_tuning_metrics}
    \begin{tabular}{lcccc}
        \toprule
        \textbf{Dataset} & \textbf{Accuracy} & \textbf{Precision} & \textbf{Recall} & \textbf{F1-Score} \\
        \midrule
        Shrec2021-manual & 91.48 & 92.57 & 91.48 & 91.50 \\
        Shrec2021-auto   & 91.48 & 91.64 & 91.48 & 91.38 \\
        MHEALTH-manual   & 99.81 & 99.81 & 99.81 & 99.81 \\
        MHEALTH-auto     & 99.81 & 99.81 & 99.81 & 99.80 \\
        UCI-HAR-manual   & 96.54 & 96.60 & 96.54 & 96.53 \\
        UCI-HAR-auto     & 97.05 & 97.08 & 97.05 & 97.04 \\
        DB4-manual       & 65.52 & 65.88 & 65.52 & 65.53 \\
        DB4-auto         & 66.86 & 67.15 & 66.86 & 66.76 \\
        Berkeley-manual  & 98.98 & 99.00 & 98.98 & 98.97 \\
        Berkeley-auto    & 98.98 & 99.00 & 98.98 & 98.97 \\
        LMDHG-manual     & 83.38 & 85.27 & 83.38 & 82.55 \\
        LMDHG-auto       & 74.25 & 77.13 & 74.25 & 72.92 \\
        \bottomrule
    \end{tabular}
\end{table}

\begin{figure}[tp]
    \centering
    \includegraphics[width=1\linewidth]{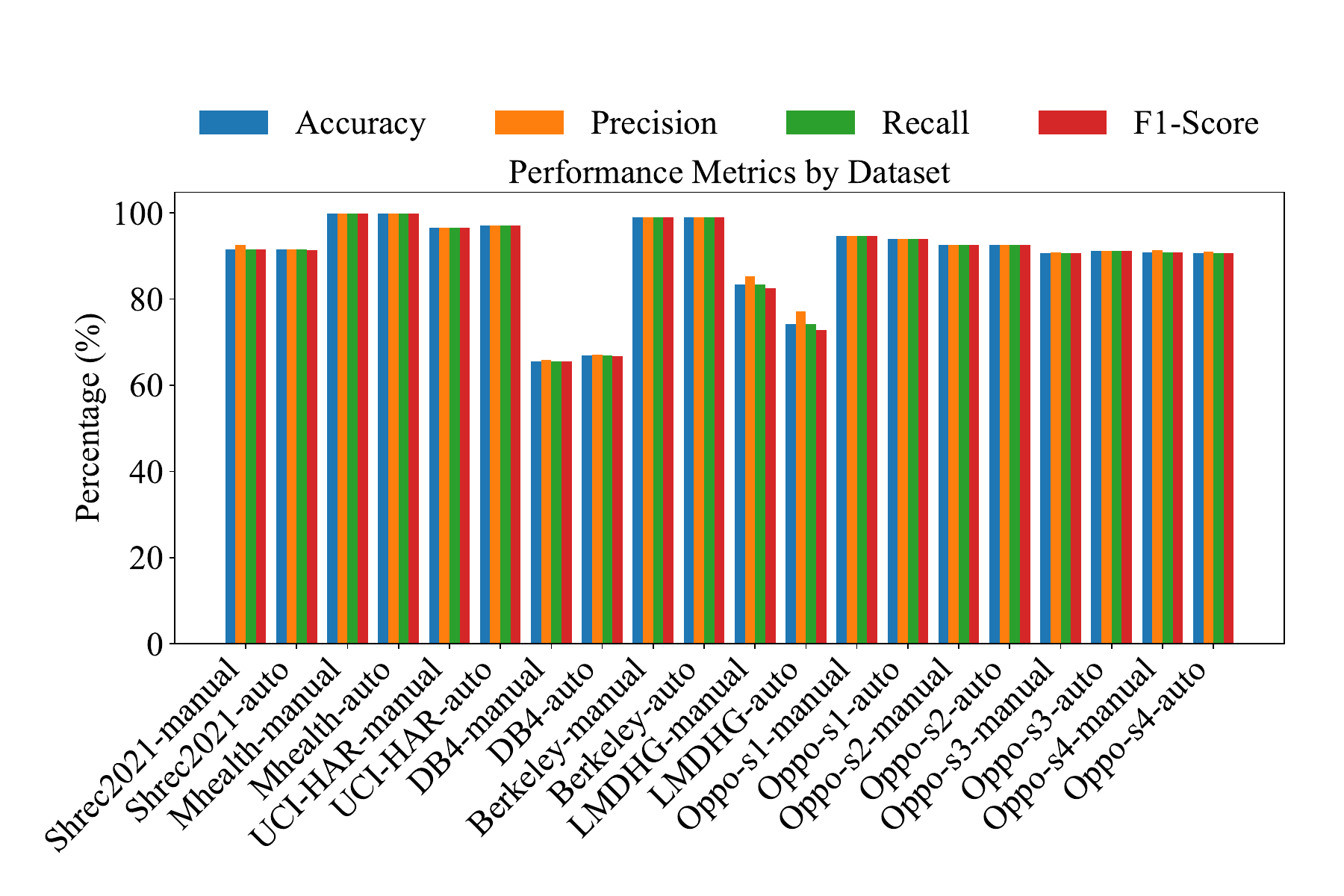}
    \caption{Ablation study comparing automatic and manual hyperparameter tuning across ten datasets. Results indicate that automatic tuning performs on par with or better than manual tuning in most cases, validating the efficiency and practicality of AutoMR’s optimization strategy.}
    \label{figure1}
\end{figure}

To evaluate the impact of automated hyperparameter tuning, an ablation study was conducted comparing automatic and manual tuning across multiple datasets. Table~\ref{table_tuning_metrics} and Figure~\ref{figure1} present a detailed comparison of model performance metrics, including accuracy, precision, recall, and F1-score. The results indicate that automated tuning achieves comparable or even slightly superior performance to manually tuned models in most datasets, suggesting that AutoMR can effectively optimize hyperparameters without requiring expert intervention. Notably, datasets such as DB4 and LMDHG show slight discrepancies, where manual tuning provides marginal improvements. This suggests that highly unstructured or sensor-specific datasets may still benefit from fine-tuned adjustments.

\section{CONCLUSION}

The evaluation of AutoMR across ten datasets demonstrates its effectiveness in automating motion recognition by streamlining data preprocessing, model training, and hyperparameter tuning. AutoMR achieves state-of-the-art performance on eight datasets, highlighting its adaptability across different sensor modalities. However, performance on DB4 and LMDHG remains lower due to the limitations of QuartzNet’s 1D convolutional architecture in capturing spatial dependencies and long-range temporal patterns. Addressing these limitations will require integrating 2D CNNs for enhanced spatial feature extraction and hybrid models that combine sequential and spatial analysis. Additionally, transformer-based architectures may further improve performance by capturing long-range dependencies in unsegmented motion sequences. Beyond model enhancements, expanding dataset coverage with diverse sensor modalities and developing adaptive hyperparameter tuning will improve generalizability. Open-source collaboration will further drive refinements, ensuring continuous advancements. 

{\small
\bibliographystyle{ieee}
\bibliography{egbib}
}

\end{document}